\documentclass[sigconf]{acmart}
\AtBeginDocument{%
  \providecommand\BibTeX{{%
    \normalfont B\kern-0.5em{\scshape i\kern-0.25em b}\kern-0.8em\TeX}}}

\setcopyright{acmlicensed}
\copyrightyear{2018}
\acmYear{2018}
\acmDOI{XXXXXXX.XXXXXXX}

\acmConference[Conference acronym 'XX]{Make sure to enter the correct
  conference title from your rights confirmation emai}{June 03--05,
  2018}{Woodstock, NY}
\acmISBN{978-1-4503-XXXX-X/18/06}

\acmSubmissionID{254}

\usepackage{times}
\usepackage{latexsym}

\usepackage[T1]{fontenc}
\usepackage{microtype}

\usepackage[utf8]{inputenc} % allow utf-8 input
\usepackage[T1]{fontenc} % use 8-bit T1 fonts
\usepackage{hyperref} % hyperlinks
\usepackage{url} % simple URL typesetting
\usepackage{booktabs} % professional-quality tables
\usepackage{amsfonts} % blackboard math symbols
\usepackage{nicefrac} % compact symbols for 1/2, etc.
\usepackage{microtype} % microtypography
\usepackage{xcolor} % colors
\usepackage{CJKutf8}
\usepackage{multirow}
\usepackage{amsmath}
\usepackage{amsthm}
\usepackage{bbding}
\usepackage{graphicx}
\usepackage{makecell}

\begin{document}

\title{Dólares or Dollars? Unraveling the Bilingual Prowess of Financial LLMs Between Spanish and English}

%\author{Jimin Huang}
\author{Xiao Zhang}
\affiliation{%
  \institution{The Fin AI}
  %\streetaddress{8600 Datapoint Drive}
  \city{Singapore}
  \country{Singapore}}
  %\postcode{78229}}
\email{xiao.zhang@thefin.ai}

\author{Ruoyu Xiang}
\affiliation{%
  \institution{The Fin AI}
  %\streetaddress{8600 Datapoint Drive}
  \city{Singapore}
  \country{Singapore}}
  %\postcode{78229}}
\email{ruoyu.xiang@thefin.ai}

\author{Chenhan Yuan}
\affiliation{%
  \institution{University of Manchester}
  %\streetaddress{8600 Datapoint Drive}
  \city{Manchester}
  \country{UK}}
  %\postcode{78229}}
\email{chenhan.yuan@postgrad.manchester.ac.uk}

\author{Duanyu Feng}
\affiliation{%
  \institution{School of Computer Science\\Sichuan University}
  % \streetaddress{P.O. Box 1212}
  \city{Chengdu}
  \state{Sichuan}
  \country{China}
}
\email{fengduanyu@stu.scu.edu.cn}

\author{Weiguang Han}
\affiliation{%
  \institution{School of Computer Science\\Wuhan University}
  \city{Wuhan}
  \state{Hubei}
  \country{China}}
  \email{han.wei.guang@whu.edu.cn}

\author{Alejandro Lopez-Lira}
\affiliation{%
  \institution{University of Florida}
  \city{Gainesville}
  \state{Florida}
  \country{USA}}
\email{alejandro.lopez-lira@warrington.ufl.edu}

\author{Xiao-Yang Liu}
\affiliation{
  \institution{Columbia University}
  \city{New York}
  \state{New York}
  \country{USA}}
\email{XL2427@columbia.edu}

\author{Sophia Ananiadou}
\affiliation{
  \institution{Department of Computer Science\\The University of Manchester}
  \city{Manchester}
  \country{UK}
}
\email{sophia.ananiadou@manchester.ac.uk}

\author{Min Peng}
\affiliation{%
  \institution{School of Computer Science\\
Wuhan University\\}
  \city{Wuhan}
  \state{Hubei}
  \country{China}}
\email{pengm@whu.edu.cn}

\author{Jimin Huang}
\affiliation{%
  \institution{The Fin AI\\}
  \city{Singapore}
  \country{Singapore}}
\email{jimin.huang@thefin.ai}

\author{Qianqian Xie}
\authornote{Corresponding Author}
\affiliation{%
  \institution{The Fin AI\\}
  \city{Singapore}
  \country{Singapore}}
\email{qianqian.xie@thefin.ai}
\renewcommand{\shortauthors}{Xiao Zhang, et al.}

\begin{abstract}
  Despite Spanish's pivotal role in the global finance industry, a pronounced gap exists in Spanish financial natural language processing (NLP) and application studies compared to English, especially in the era of large language models (LLMs). To bridge this gap, we unveil Toisón de Oro, the first bilingual framework that establishes instruction datasets, finetuned LLMs, and evaluation benchmark for financial LLMs in Spanish joint with English. We construct a rigorously curated bilingual instruction dataset including over 144K Spanish and English samples from 15 datasets covering 7 tasks. Harnessing this, we introduce FinMA-ES, an LLM designed for bilingual financial applications. We evaluate our model and existing LLMs using FLARE-ES, the first comprehensive bilingual evaluation benchmark with 21 datasets covering 9 tasks. The FLARE-ES benchmark results reveal a significant multilingual performance gap and bias in existing LLMs. FinMA-ES models surpass SOTA LLMs such as GPT-4 in Spanish financial tasks, due to strategic instruction tuning and leveraging data from diverse linguistic resources, highlighting the positive impact of cross-linguistic transfer. All our datasets, models, and benchmarks have been released\footnote{https://github.com/chancefocus/PIXIU}.
\end{abstract}






\maketitle

\section{Introduction}
    % spanish datasets：6 datasets， pieces of data， 3.6MB
    In light of digital transformations in the finance sector, recognizing the significance of global languages becomes paramount. A salient observation here is the substantial number of Spanish speakers. There are 485 million native Spanish speakers globally, asserting its position as the fourth most spoken language\cite{julian2020most}. The U.S., a hub for fintech innovations, hosts 15 million individuals who speak Spanish as a secondary language. Additionally, the expanding digital user base necessitates that financial platforms consider such linguistic demographics, especially with Spanish speakers poised for significant growth by 2030~\cite{hoff2017language}.

    The integration of artificial intelligence (AI) in financial technologies (FinTech) has significantly accelerated advancements in the sector, particularly through the application of pre-trained language models (PLMs)~\cite{devlin-etal-2019-bert} and recent large language models (LLMs)~\citep{openai2023gpt4,touvron2023llama} in natural language processing (NLP).
    These models, trained on vast amounts of text data, have the potential to understand, interpret, and generate human-like text, thereby transforming the way financial information is analyzed and processed~\cite{wei2022emergent}.
    They have been pivotal in transforming financial services, enabling sophisticated capabilities ranging from stock price forecasting to comprehensive financial analytics~\cite{xie2023pixiu,wu2023bloomberggpt,lopez2023can,li2023chatgpt}. 

    Despite these advancements, a notable language disparity persists in the realm of FinTech~\cite{araci2019finbert,han2023select,xie2023wall,lopez2023can,li2023chatgpt}. 
    The development and application of financial PLMs, including models like FinBERT~\cite{araci2019finbert}, FLANG~\cite{shah2022flue}, and BloombergGPT~\cite{shah2022flue}, have predominantly concentrated on English. 
    %Financial pre-trained language models (PLMs) like finBERT~\cite{araci2019finbert} and FLANG~\cite{shah2022flue} cater to financial lexicons but with limited cross-lingual adaptability. T
    The recent financial LLM BloombergGPT~\cite{wu2023bloomberggpt}, FinGPT~\cite{wang2023fingpt} and PIXIU~\cite{xie2023pixiu}, continue this English-centric trend. 
    Although there are attempts to adapt financial LLMs to other languages, such as Chinese, through models like DISC-FinLLM~\cite{chen2023discfinllm} and CFGPT~\cite{li2023cfgpt}, a significant gap remains in the development of Spanish-focused financial LLMs.

%  \begin{figure}[h]
%   \centering
%   \includegraphics[width=\linewidth]{FinMA}
%   \caption{An overview of multi-task and multi-modal instruction tuning of FinMA for diverse financial tasks.}
%   %\Description{A woman and a girl in white dresses sit in an open car.}
%   \label{fig:FinMA}
% \end{figure}

\begin{CJK}{UTF8}{gkai}
	To address it, we propose Toisón de Oro, a novel bilingual framework meticulously designed to bridge the research gap. "Toisón de Oro" includes the first open-source multi-task and Spanish-English bilingual instructional data (FIT-ES) with 151k samples from 15 datasets covering 7 tasks, the first open-source Spanish-English bilingual evaluation benchmark (FLARE-ES) with 21 datasets covering 9 tasks, and the pioneering open-source Spanish-English bilingual financial LLM, FinMA-ES. This model is derived by finetuning LLaMA2 7B model~\citep{touvron2023llama} using FIT-ES. The contributions of "Toisón de Oro" are manifold: 1) We provide open access to all components, promoting widespread adoption and further innovation in the financial NLP sector. 2) Our bilingual solution addresses the critical need for multilingual financial analysis, enabling global firms to navigate and analyze diverse data sets effectively. 3) The scalable multi-task framework of "Toisón de Oro" supports a broad spectrum of financial applications, facilitating a unified and efficient approach to financial analytics. 

To build the multi-task and bilingual instruction data, we sourced from Spanish and English financial tasks such as sentiment analysis, news headline classification, annual report text summarization, and examination question answering. This led to the creation of Financial Instruction Tuning Data in Spanish and English (FIT-ES), combining task-specific Spanish instructions with the respective data samples. 
In pursuit of enhancing bilingual financial analytics, we propose the bilingual financial LLM, FinMA-ES by finetuning LLaMA2 7B with FIT-ES. 
For model evaluation, we further build the bilingual FLARE-ES benchmark by including validationa and test set from FIT-ES, and extra 6 unseen datasets and 2 unseen tasks. This deliberate inclusion is aimed at evaluating the model's generalization capabilities across a diverse array of financial contexts and tasks.

We evaluate the performance of FinMA and exisiting SOTA LLMs with FLARE-ES.
Our experimental analysis revealed three critical insights: 1) Existing LLMs, including GPT-4, demonstrate a pronounced performance gap in Spanish financial tasks, highlighting a significant disparity in their effectiveness across languages. 2) The FinMA-ES models excel in bilingual financial analysis, surpassing established models like GPT-4 in key Spanish financial tasks. This superiority underscores the pivotal role of instruction tuning, employing both target language and high-resource language datasets to enhance model capabilities. 3) Interestingly, fine-tuning LLMs with data from low-resource languages not only addresses gaps in those languages but also unexpectedly boosts the models' performance in high-resource language datasets, suggesting a beneficial cross-linguistic transfer effect.
		
Our contributions can be summarized as follows: 1) We created the first bilingual framework specifically designed for Spanish-English financial NLP and prediction tasks.  
2) We developed the first bilingual and multi-task instruction tuning data.
3) We developed and fine-tuned the FinMA-ES model, the first LLM optimized for processing and understanding bilingual financial data on both Spanish and English.
		4) We established the FLARE-ES benchmark, the first open-source comprehensive set of evaluations that allows for the cross-lingual assessment of models on both Spanish financial tasks and English financial tasks.
  5) Our FLARE-ES benchmark evaluation indicates that FinMA-ES models notably outperform leading LLMs like GPT-4 in Spanish financial tasks due to strategic instruction tuning and data from both low- and high-resource languages, revealing a critical multilingual performance disparity and the unexpected benefits of cross-linguistic transfer.
		
\end{CJK}

\section{Related Work}
\label{sec:related}
	\textbf{Financial Language Models}
    There is a pronounced absence of models, both PLMs and LLMs, specifically designed for Spanish or bilingual Spanish-English applications. Models such as finBERT \citep{araci2019finbert} and FLANG \citep{shah2022flue} excel in processing English financial texts but offer limited utility in cross-lingual scenarios. Similarly, the advanced BloombergGPT \citep{wu2023bloomberggpt}, despite its massive scale, perpetuates this English-centric approach. Other recent developments, like FinGPT \citep{wang2023fingpt}, InvestLM \citep{yang2023investlm}, and PIXIU \citep{xie2023pixiu}, continue to focus predominantly on English. While there have been efforts to create financial LLMs for other languages, such as Chinese, with models like DISC-FinLLM \citep{chen2023discfinllm} and CFGPT \citep{li2023cfgpt}, Spanish remains notably underserved. This significant gap in Spanish and bilingual financial language models highlights the unique importance and potential impact of our work in developing bilingual financial LLMs, aiming to bridge this linguistic divide and cater to a wider, more diverse audience in the financial domain.
 
    \textbf{Financial Evaluation Benchmark}
    In the sphere of financial NLP, there has been a significant focus on developing benchmarks for English and Chinese. \cite{shah2022flue} introduced the FLUE benchmark, offering a diverse set of financial NLP tasks in English. Complementing this, \cite{xie2023pixiu} developed FLARE, another English benchmark for evaluating financial LLMs with a broad range of tasks. \cite{wang2023fingpt} also proposed an English benchmark for financial LLMs. In the Chinese context, the BBT-CFLEB benchmark by \cite{lu2023bbt} and FinEval by \cite{zhang2023fineval} have been instrumental in advancing Chinese financial NLP. Further contributing to this are DISC-FinLLM by \cite{chen2023discfinllm}, which offer unique perspectives and tasks for evaluating Chinese financial language models. Despite these strides in English and Chinese benchmarks, the absence of comprehensive Spanish financial NLP benchmarks is evident, underscoring a significant gap in the field.

    \textbf{Open Sourced Large Language Models}
    In the current landscape of AI democratization, while general models like LLaMA \citep{touvron2023llama} and its instruction-following variants, Alpaca \cite{alpaca}, and Vicuna-13B \cite{vicuna2023}, have shown significant progress, the development in financial LLMs exhibits a language bias. English financial LLMs have been developed, as seen in PIXIU~\cite{xie2023pixiu} and FinGPT \cite{wang2023fingpt}, and Chinese financial LLMs have advanced with DISC-FinLLM \cite{chen2023discfinllm}, CFGPT~\cite{li2023cfgpt}, and InvestLM \citep{yang2023investlm}. Lince-zero \cite{lince-zero} represents progress for Spanish LLMs. However, this progression still leaves a notable gap in open-sourced Spanish and bilingual financial LLMs, underscoring a critical area for the global financial industry.
\section{Method}
\subsection{FIT-ES: Financial Instruction Tuning Dataset-Encompassing Spanish}
In this section, we present the composition and development of our Financial Instruction Tuning dataset, FIT-ES, which is the foundation for our Spanish financial LLM. We detail the origins of the raw data, enumerate the specific tasks encompassed within FIT-ES, and describe the meticulous process employed to construct the dataset from this raw data. Unique among existing resources, FIT-ES distinguishes itself as the first instruction-tuning dataset specifically crafted for Spanish financial LLMs.

\subsubsection{Raw Data}
	\label{sec:raw-data}

 Our Spanish instruction tuning dataset, is developed from publicly available sources, encompassing a range of datasets for financial NLP tasks and examination content. This dataset is rooted in authentic finance-related scenarios and benefits from the high-quality annotations typically provided by domain experts in open-sourced data. As shown in Table~\ref{tab:raw-data}, this dataset includes 15 datasets for 7 financial NLP and prediction tasks in both Spanish and English. 
 This bilingual data selection strategy aims to enhance the cross-lingual capabilities of our financial language models and ensure a balanced representation of both languages in the training process.

 \begin{table*}[htb!]
		\centering
		\scriptsize
		\caption{The details of the raw data and instruction data.}
		\label{tab:raw-data}
        \resizebox{0.999\textwidth}{!}{
		\begin{tabular}{lllllllll}
			\toprule
			\textbf{Data}&\textbf{Task}&\textbf{Language}&\textbf{Raw}&\textbf{Instruction}&\textbf{Data Types}&\textbf{Modalities}&\textbf{License}\\
			\midrule
            MultiFin~\citep{jorgensen-etal-2023-multifin}&classification&Spanish&2,066&2,066&article headlines&text&MIT License\\

            FNS-2023\footnote{https://wp.lancs.ac.uk/cfie/fns2023/}&text summrization&Spanish&232&232&annual reports&text&Public\\

             EFP\footnote{https://efpa-eu.org/}&question answering&Spanish&37&37&exam questions&text&Public\\

             EFPA~\footnote{https://efpa-eu.org/}&question answering&Spanish&228&228&exam questions&text&Public\\

             TSA~\citep{pan2023evaluation}&sentiment analysis&Spanish&3,892&3,892&news headlines&text&Public\\

             FinanceES~\citep{FinanceES}&sentiment analysis&Spanish&7,980&7,980&news headlines&text&Public\\
            FPB&sentiment analysis&English&4,845&48,450&news&text&CC BY-SA 3.0\\
            
            FiQA-SA~\citep{maia201818}&sentiment analysis&English&1,173&11,730&news headlines, tweets&text&Public\\
            
            Headline~\citep{sinha2021impact}&news headline classification&English&11,412&11,412&news headlines&text&CC BY-SA 3.0\\
            
            NER~\citep{alvarado2015domain}&named entity recognition&English&1,366&13,660&financial agreements&text&CC BY-SA 3.0\\

            FinQA~\citep{chen2021finqa}&question answering&English&8,281&8,281&earnings reports&text, table&MIT License\\

            ConvFinQA~\citep{chen2022convfinqa}&question answering&English&3,892&3,892&earnings reports&text, table&MIT License\\

            BigData22~\citep{soun2022accurate}&stock movement prediction&English&7,164&7,164&tweets, historical prices&text, time series& Public\\

            ACL18~\citep{xu2018stock}&stock movement prediction&English&27,053&27,053&tweets, historical prices&text, time series& MIT License\\

            CIKM18~\citep{wu2018hybrid}&stock movement prediction&English&4,967&4,967&tweets, historical prices&text, time series& Public\\
			\bottomrule
		\end{tabular}}
	\end{table*}
 
 \textbf{Classification.} The task integrates two distinct datasets, MultiFin~\citep{jorgensen-etal-2023-multifin} and Gold news headline~\citep{sinha2021impact}, to assess the model's classification prowess across both Spanish and English financial texts. The MultiFin dataset, focusing on Spanish headlines, compiles 2,066 articles spanning six critical financial categories: "Business \& Management", "Finance", "Government \& Controls," "Industry," "Tax \& Accounting," and "Technology," challenging the model to accurately categorize each headline into its respective sector. Concurrently, the Gold news headline dataset delves into English financial texts concerning gold, encompassing a period from 2000 to 2019. It features a detailed classification scheme with nine tags: “price or not,” “price up,” “price down,” “price stable,” “past price,” “future price,” “past general,” “future general,” and “asset comparison,” aimed at dissecting the headlines into binary classifications based on their implied price movement or market sentiment. This comprehensive classification task not only tests the model's linguistic flexibility and sector-specific knowledge across two languages but also its ability to discern and predict market trends from textual data, reflecting its applicability in automated financial news analysis.

\textbf{Question Answering.} Our investigation extends into the domain of question answering (QA), a critical task for evaluating the model's comprehension and application of financial knowledge across both Spanish and English datasets. In the Spanish context, we utilize the EFP and EFPA datasets\footnote{https://efpa-eu.org/}, which consist of questions derived from financial examinations provided by official examiner associations. The EFP dataset challenges the model with 37 questions, each offering three possible answers ("A," "B," or "C"), thereby testing the model's proficiency in accurately identifying the correct response based on the given financial scenario. The EFPA dataset further expands this challenge, presenting 228 questions with four answer choices ("A," "B," "C," or "D"), encompassing a wider array of financial topics, including economic knowledge, fundamental financial concepts, and detailed computations related to financial products.
Transitioning to the English datasets, we employ FinQA~\citep{chen2021finqa} and ConvFinQA~\citep{chen2022convfinqa} to assess the model's QA capabilities in a different linguistic and financial context. The FinQA dataset, comprising 8,281 questions and answers extracted from earnings reports, necessitates the model to navigate through complex text and table data to derive accurate answers, reflecting its ability to handle multifaceted financial documents. Similarly, the ConvFinQA dataset, with 3,892 questions and answers based on earnings reports, challenges the model to understand and respond to queries within a conversational context, highlighting its capacity for nuanced language understanding and information retrieval within financial discussions.
%Together, these datasets provide a comprehensive framework for assessing the model's question-answering abilities across languages, emphasizing the importance of domain-specific knowledge, linguistic versatility, and analytical prowess in interpreting and responding to financial inquiries.

\textbf{Text Summarization.} The task of text summarization within our study focuses on condensing voluminous financial documents into concise, informative abstracts, a critical capability for enhancing the accessibility and usability of financial information. This task employs the FNS-2023 dataset\footnote{https://wp.lancs.ac.uk/cfie/fns2023/}, which comprises a collection of 232 annual reports from various financial companies. These reports, rich in detailed financial data and narratives, present a unique challenge: to distill the essence of each document into a summary that captures the most crucial information while maintaining the factual integrity and coherence of the original text. %With 50 reports designated for testing and an additional 146 for training, the task assesses the model's ability to identify key points, extract relevant information, and generate summaries that provide stakeholders with a clear, concise overview of a company's financial health and performance.

\textbf{Financial Sentiment Analysis.} This task delves into the intricate sentiment dynamics within financial texts, employing the TSA~\citep{pan2023evaluation} and FinanceES~\citep{FinanceES} datasets for Spanish sentiment analysis, and the FPB and FiQA-SA datasets for English. The TSA dataset, comprising 3,892 entries from financial news and tweets, is meticulously annotated to reflect sentiments as positive, negative, or neutral, providing a nuanced spectrum of market emotions in Spanish. Similarly, the FinanceES dataset enriches this analysis with 7,980 Spanish financial news headlines, each labeled to indicate the underlying sentiment, thus offering a comprehensive base for evaluating the model's sentiment detection accuracy in Spanish financial discourse. Transitioning to English, the FPB~\citep{malo2014good} dataset introduces an expansive collection of 4,845 news items, with sentiment labels that challenge the model's ability to discern and classify sentiments in English financial news. Complementing this, the FiQA-SA~\citep{maia201818} dataset includes 1,173 entries combining news headlines and tweets, each with sentiment annotations, further broadening the scope of sentiment analysis in English financial texts. This dataset not only tests the model's semantic understanding but also its capacity to navigate the subtleties of sentiment expression in diverse formats, from concise tweets to more detailed news articles. 
%The financial sentiment analysis task, through these datasets, aims to quantify the model's adeptness at identifying and categorizing sentiment across a vast array of financial texts in both Spanish and English. 

\textbf{English-Only Financial Tasks.} Our study includes English-specific tasks leveraging datasets for named entity recognition (NER)~\citep{alvarado2015domain}, and stock movement prediction to evaluate the model's financial analysis capabilities. The NER task uses a dataset of 1,366 financial agreements to test entity identification within financial texts. For stock movement prediction, three datasets—BigData22~\citep{soun2022accurate} with 7,164 entries, ACL18~\citep{xu2018stock} featuring 27,053 entries, and CIKM18~\citep{wu2018hybrid} comprising 4,967 entries—challenge the model to forecast stock prices based on textual and quantitative data. These tasks assess the model's proficiency in extracting critical information and predicting market trends, showcasing its utility in financial analytics and decision-making processes within the English financial domain.

\subsubsection{Instruction Construction}
    % instruction before, general prompt, sepcial adjustment and explanation for EFP and EFPA
We crafted financial instruction datasets from the raw data outlined in Table \ref{tab:raw-data}, with carefully designed instructions by domain experts who are proficient in both Spanish and English\footnote{For detailed instruction, please see Appendix \ref{sec:ins}}. The construction of instruction tuning samples follows a general structured template: 

\colorbox[gray]{0.95}{Instruction: [task prompt]\quad Text: [input] \quad Response: [output]}\\

which integrates human-designed instructions with input texts and their corresponding outputs. [task prompt] is the prompt designed for each data, [input text] is the input financial data from each data, e.g. the historical prices and tweets or headlines, [output] is the corresponding output for input text, e.g. sentiment label of input text from ["Positive", "Negative", "Neutral"] and ["positivo", "negativo", "neutral"] in Spanish. 

MultiFin, FNS-2023, FinanceES, and TSA datasets adopted a unified approach due to their similar task structures and data types. However, the EFP and EFPA datasets required two distinct sets of instructions Customized to their respective answer choices.

%Tailored prompts were developed for the EFP and EFPA datasets to address their unique formats, resulting in two distinct sets of instructions that cater to their respective answer choices. A unified approach was applied to the FinanceES and FinanceES-TSA 2022 datasets due to their similar task structures and data types, utilizing a single prompt to ensure consistency in response accuracy.
    
    %We developed the instructions for all datasets and have made them as detailed as possible to ensure no ambiguity. Since EFP and EFPA have the same types of input data, and EFP consists of three answers, and EFPA consists of four answers, we have created two prompts for the two datasets to ensure the accuracy of the responses. We use the same prompt for FinanceES and FinanceES-TSA 2022, since they have the same data types of input data and task formulation, also the same answer choices. 

    %We ask domain experts to write 10 diverse instructions for all datasets except the ConvFinQA, where we only use one instruction.

    %We show the instruction examples in Table \ref{tab:prompt}. 

\subsection{FinMA-ES: Financial Large Language Model in Both English and Spanish}
\label{sec:finma-es}
We propose the \textbf{FinMA-ES-Bilingual}, a bilingual financial large language model, through fine-tuning the LLaMA2-7B backbone model~\cite{touvron2023llama}, specifically aimed at enhancing performance in both English and Spanish financial tasks based on FIT-ES. 
This fine-tuning involved 5 epochs using the AdamW optimizer~\cite{kingma2014adam}, characterized by a batch size of 1, a learning rate of 3e-4, and a weight decay set to 1e-5. The entire process was executed on a robust computational framework provided by 2 NVIDIA HGX A100 SXM4 80GB GPUs. %, ensuring efficient and effective optimization.
 We also proposed the \textbf{FinMA-ES-Spanish} which is only finetuned with the Spanish data, for conducting the ablation study. 
 This additional analysis aims to evaluate the contribution of Spanish training data towarding the overall model performance, highlighting the value of language-specific data in enhancing FinMA-ES's bilingual capabilities. %This strategic inclusion of a monolingual ablation study serves to deepen our understanding of the model's linguistic versatility and adaptability within the financial NLP domain.

\subsection{FLARE-ES: Financial Evaluation Benchmark on Spanish and English}
\label{sec:benchmark}
\begin{figure}[h]
  \centering
  \includegraphics[width=\linewidth]{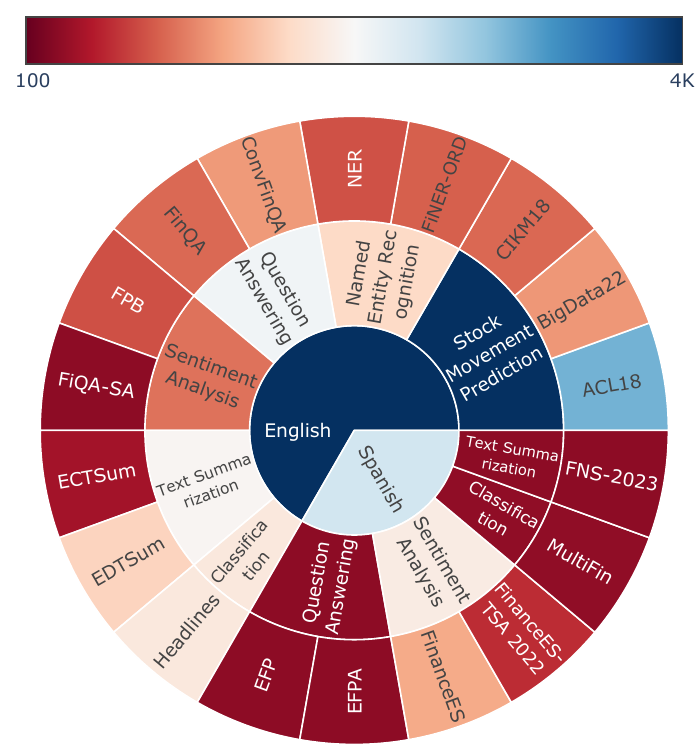}
  \caption{An overview of bilingual benchmark FLARE-ES.}
  %\Description{A woman and a girl in white dresses sit in an open car.}
  \label{fig:flare}
\end{figure}

We further propose the FLARE-ES evaluation benchmark with 21 datasets from 11 tasks, to holistically evaluate the capabilities of LLMs in the financial domain, covering both English and Spanish languages. Each task is designed to probe different aspects of financial data understanding and generation, and the financial prediction, utilizing specific metrics for a detailed assessment, as shown in Table \ref{tab:eval} and Figure~\ref{fig:flare}.

\begin{table*}[htb!]
		\centering
		\scriptsize
		\caption{The details of our evaluation datasets and evaluation metrics.
        In order to compare performance across different models, such as GPT-4 and ChatGPT, we maintain consistency by using identical datasets with the same data distributions across all models during training.
        %To compare the performance with different models such as GPT-4, and ChatGPT, we keep the same numbers and data distributions of our test datasets for every dataset. 
        }
		\label{tab:eval}
            \resizebox{0.999\textwidth}{!}{
		\begin{tabular}{lllccc}
			\toprule
                Data&Task&Language&Valid&Test&Evaluation \\
			%\textbf{Data}&\textbf{Task}&\textbf{Valid}&\textbf{Test}&\textbf{Evaluation}\\
			\midrule
   
			MultiFin~\citep{jorgensen-etal-2023-multifin}&classification&Spanish&230&368&F1, Accuracy\\
			
			FNS-2023\footnote{https://wp.lancs.ac.uk/cfie/fns2023/}
			&text summarization&Spanish
			&36
			&50
			&rouge1, rouge2, rougeL\\
			
			EFP\footnote{https://efpa-eu.org/}
			&question answering&Spanish
			&5
			&210
			&F1, Accuracy\\
			
			EFPA\footnote{https://efpa-eu.org/}
			&question answering&Spanish
			&35
			&50
			&F1, Accuracy\\
			
			TSA~\citep{pan2023evaluation}
			&sentiment analysis&Spanish
			&200
			&726
			&F1, Accuracy\\
			
			FinanceES~\citep{FinanceES}
			&sentiment analysis&Spanish
			&1,272
			&1,621
			&F1, Accuracy\\

			FPB~\citep{malo2014good}
			&sentiment analysis&English
			&\textcolor{black}{775}
			&970
			&F1, Accuracy\\
			
			FiQA-SA~\citep{maia201818}
			&sentiment analysis&English
			&188
			&235
			&F1\\
			
			\textcolor{black}{Headlines}~\citep{sinha2021impact}
			&news headline classification&English
			&\textcolor{black}{1,141}
			&2,283
			&Avg F1\\
			
			\textcolor{black}{NER}~\citep{alvarado2015domain}
			&named entity recognition&English
			&103
			&980
			&Entity F1\\
			
			FinQA~\citep{chen2021finqa}
			&question answering&English
			&883
			&1,147
			&EM Accuracy\\

			ConvFinQA~\citep{chen2022convfinqa}
			&question answering&English
			&2,210
			&1,490
			&EM Accuracy\\
			
			BigData22~\citep{soun2022accurate}
			&stock movement prediction&English
			&798
			&1,470
			&\textcolor{black}{Accuracy}, MCC\\
			
			ACL18~\citep{xu2018stock}
			&stock movement prediction&English
			&2,560
			&3,720
			&\textcolor{black}{Accuracy}, MCC\\
			
			CIKM18~\citep{wu2018hybrid}
			&stock movement prediction&English
			&431
			&1,140
			&\textcolor{black}{Accuracy}, MCC\\\hline
            \textcolor{black}{FiNER-ORD}~\citep{shah2023finer}
			&\textcolor{black}{named entity recognition}&English
			&-
			&1080
			&\textcolor{black}{Entity F1}\\
   
			\textcolor{black}{ECTSum~\cite{mukherjee2022ectsum}}
			&\textcolor{black}{text summarization}&English
			&-
			&495
			&\textcolor{black}{ROUGE, BERTScore, BARTScore}\\
			
			\textcolor{black}{EDTSum~\cite{zhou2021trade}}
			&\textcolor{black}{text summarization}&English
			&-
			&2000
			&\textcolor{black}{ROUGE, BERTScore, BARTScore}\\
			
			\textcolor{black}{German~\cite{misc144}}
			&\textcolor{black}{credit scoring}&English
			&-
			&1000
			&\textcolor{black}{F1, MCC}\\
			
			\textcolor{black}{Australian~\cite{misc143}}
			&\textcolor{black}{credit scoring}&English
			&-
			&690
			&\textcolor{black}{F1, MCC}\\
			
			\textcolor{black}{FOMC~\cite{shah2023trillion}}
			&\textcolor{black}{hawkish-dovish classification}&English
			&-
			&496
			&\textcolor{black}{F1, Accuracy}\\
			
                \bottomrule
		\end{tabular}}
	\end{table*}
 \subsubsection{Evaluation Tasks and Datasets}
To thoroughly assess models' performance, we incorporate datasets from the same sources used in training, as well as additional 6 datasets and 2 tasks not employed during the training phase. This approach is designed to rigorously evaluate the generalization capabilities of various models in financial tasks.

 \textbf{1) Sentiment analysis}. We utilize the TSA and FinanceES~\citep{FinanceES} datasets for Spanish sentiment analysis, alongside the FPB~\citep{malo2014good} and FiQA-SA~\citep{maia201818} datasets for English, for evaluating models' abilities to discern and categorize sentiments as positive, negative, or neutral. Following previous works~\citep{xu2018stock,xie2023wall}, performance across these datasets is meticulously quantified using Accuracy (ACC) and the F1 Score. 
 %Accuracy provides a direct measure of the models' overall effectiveness in sentiment classification, while the F1 Score, as the harmonic mean of precision and recall, offers a balanced assessment of models' sensitivity and specificity in detecting sentiments. % We need the paper that used the ACC and F1 score 
 
\textbf{2) Classification}. We leverage the MultiFin dataset for Spanish, containing 230 validation and 368 test samples, alongside the English Headlines dataset, which includes 1,141 validation and 2,283 test samples. These datasets challenge LLMs to accurately classify financial news articles into predefined categories reflecting the model's understanding of domain-specific content. Performance is evaluated using the F1 Score and Accuracy metrics. 
%The F1 Score, calculated as the harmonic mean of precision and recall, is particularly valuable in assessing the model's balanced performance across different classes, essential in financial datasets where class distribution can be skewed. Accuracy, the proportion of correctly predicted observations to the total observations, offers a straightforward metric of the model's overall classification capability.

\textbf{3) Text sumarization}. This task utilizes the Spanish FNS-2023 dataset\footnote{https://wp.lancs.ac.uk/cfie/fns2023/}, consisting of 36 validation and 50 test samples, alongside English datasets including ECTSum and EDTSum derived from additional sources not utilized in training.
To quantitatively measure the quality of the model-generated summaries against reference summaries,
we employ ROUGE scores (Recall-Oriented Understudy for Gisting Evaluation), BERTScore, an automated evaluation metric for text generation, and BARTScore, which is a superior text generation evaluation metric, excelling in 16 of 22 tests,
with accessible code and interactive leaderboard for comprehensive assessment.).
ROUGE metrics provide a comprehensive analysis of summarization performance by measuring unigram overlap (rouge1), bigram overlap (rouge2), and the longest common subsequence (rougeL). BERTScore, utilizing BERT's contextual embeddings, further refines this evaluation by comparing the semantic similarity between generated and reference texts, allowing for a nuanced understanding of content quality beyond mere textual overlap.
%which can evaluates text generation quality effectively across diverse NLP tasks,  ROUGE metrics, including rouge1 (measuring the overlap of unigrams), rouge2 (measuring the overlap of bigrams), and rougeL (considering the longest common subsequence), provide a multi-dimensional view of summarization performance. Additionally, BERTScore, a metric leveraging the contextual embeddings from BERT, offers a nuanced evaluation by comparing the semantic similarity between the generated and reference texts. 
%Unlike ROUGE, which is based on discrete token overlap, BERTScore assesses the quality of summarization from a semantic similarity standpoint, making it highly effective in recognizing paraphrased content that retains the original meaning without necessarily replicating the exact wording.
% the index that appears first should be stated and introduced.

\textbf{4) Question answering}. This task utilizes the EFP and EFPA datasets for Spanish, featuring 5 and 35 validation samples, and 210 and 50 test samples, respectively, along with English datasets FinQA and ConvFinQA. This task assesses models on their ability to retrieve or generate accurate answers to financial questions based on provided information. Performance is evaluated through F1 Score and Accuracy for Spanish datasets~\citep{xu2018stock,xie2023wall}. For English datasets, EM (Exact Match) Accuracy is used, focusing on the model's ability to produce answers that exactly match the gold standard responses.

\textbf{5) Named entity recognition}. This task is to identify essential entities in the financial domain, namely individuals, companies, and geographic locations. These entities play a crucial role in constructing comprehensive financial knowledge graphs and we utilized the FIN dataset~\citep{alvarado2015domain}, which comprises sentences extracted from publicly available financial agreements found in U.S. Security and Exchange Commission (SEC) filings. Additionally, we have manually annotated the entity types, categorizing them into LOCATION (LOC), ORGANISATION (ORG), and PERSON (PER) classifications.

\textbf{6) Credit scoring}. This task focuses on evaluating performance using specific datasets related to German and Australian Credit Scoring. It aims to assess how well models perform in predicting credit scores based on these datasets, employing metrics such as F1 Score and Matthews Correlation Coefficient (MCC)~\citep{xie2023wall} for evaluation.

\textbf{7) Hawkish-dovish classification}. The Hawkish-Dovish classification distinguishes sentences in financial texts as "hawkish" or "dovish", requiring intricate comprehension of monetary policy language beyond conventional sentiment analysis. This approach utilizes the FOMC dataset~\citep{shah2023trillion}, where Federal Open Market Committee (FOMC) meeting sentences are carefully annotated to reflect their economic stance. Similary to sentiment analysis, we use F1 and Accuracy metrics for evaluation.

\section{Experiments}
\label{sec:experiments}
The proposed FIT-ES and FLARE-ES allow us to train, select the model, and evaluate the performance of LLMs on financial understanding and predictions. This section investigates how powerful our fine-tuned models and other LLMs are on FLARE-ES.
We compare FinMA-ES with the following baselines: %1) BloombergGPT~\citep{wu2023bloomberggpt}. The only large language model with 50B parameters pre-trained with the financial texts.
    \begin{enumerate}
        \item GPT-4~\citep{openai2023gpt4}. A powerful instruction following large language model with around 1T parameters proposed by OpenAI.
        \item ChatGPT\footnote{https://openai.com/blog/chatgpt}. An instruction following large language model with 175B parameters from OpenAI.
        \item LLaMA2-7B~\citep{touvron2023llama}. An open-sourced large language model by META with 7B parameters.
        \item Falcon-7B\footnote{https://huggingface.co/tiiuae/falcon-7b}. A causal decoder-only model built by TII with 7B parameters.
        \item Bloomz-7B1-mt~\citep{muennighoff2022crosslingual}. An open-access multilingual large language model with 7B parameters.
        \item Lince-zero\footnote{https://huggingface.co/clibrain/lince-zero}. An open-sourced Spanish-instruction tuned large language model based on Falcon-7B.
        \item FinMA-7B-full~\citep{xie2023pixiu}. An open-sourced English financial instruction tuned large language model based on LLaMA2 with 7B parameters.
        \item FinMA-30B-nlp~\citep{xie2023pixiu}. An open-sourced English financial large language model fine-tuned using only financial NLP tasks based on LLaMA2 with 30B parameters.
        \end{enumerate}

\begin{figure}[h]
  \centering
  \includegraphics[width=\linewidth]{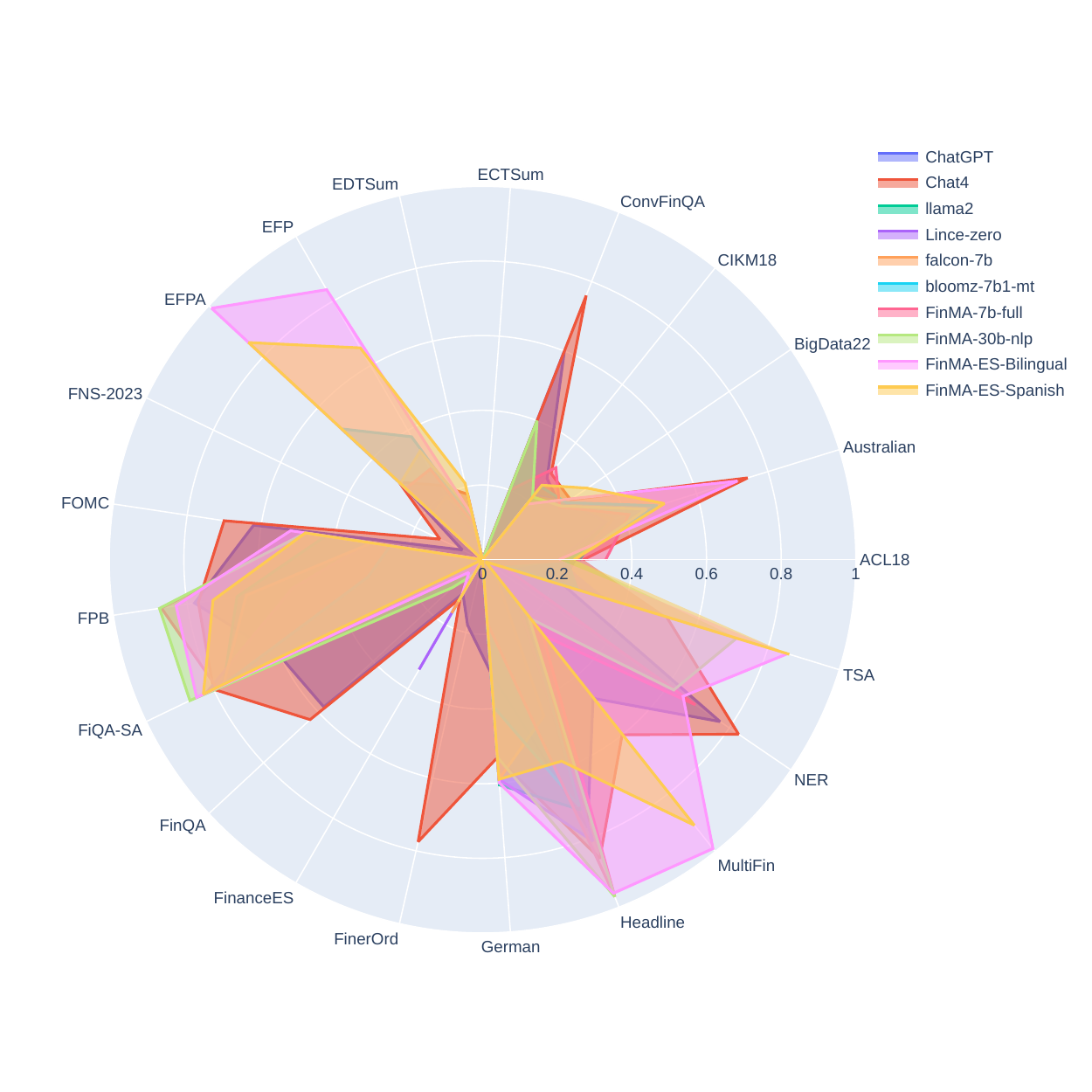}
  \caption{The radar graph of the performance for all methods on FLARE-ES.}
  %\Description{A woman and a girl in white dresses sit in an open car.}
  \label{fig:radar}
\end{figure}

\begin{table*}[htb!]
    \centering
    \scriptsize
	\caption{The performance of different LLMs on the FLARE-ES benchmark. Results of ChatGPT, GPT-4, LLaMA2, Lince-zero, FinMA-7B-full, Falcon-7B, Bloomz-7B1-mt,  FinMA-7B-full, and FinMA-30B-nlp. Test datasets were built to be the same data on every LLM.}
		\label{tab:per}
  
    \setlength{\extrarowheight}{2pt}	
    \resizebox{\textwidth}{!}{
    \begin{tabular}{llccccccccccccc}
    %{@{}l@{\hspace{4pt}}l@{\hspace{4pt}}c@{\hspace{4pt}}c@{\hspace{4pt}}c@{\hspace{4pt}}c@{\hspace{4pt}}c@{\hspace{4pt}}c@{\hspace{4pt}}c@{\hspace{4pt}}c@{\hspace{4pt}}c@{\hspace{4pt}}c@{\hspace{4pt}}c@{\hspace{4pt}}c@{}}

     %\textbf{Dataset} & \textbf{Metrics} & \makecell{\textbf{Chat}\\\textbf{GPT}}&\makecell{\textbf{Chat}\\\textbf{4}}&\textbf{llama2} &\makecell{\textbf{Lince-}\\ \textbf{zero}}&\textbf{falcon-7b} & \textbf{bloomz-7b1-mt} & \textbf{FinMA-7b-full} & \textbf{FinMA-7B-n} &\textbf{FinMA-7B-t} & \textbf{Finma-30B-n} &\textbf{FinMA-ES-Bilingual} & \textbf{FinMA-ES-Spanish}
    
    \toprule

    \textbf{Dataset} & \textbf{Metrics} & 
    \makecell{\textbf{Chat}\\\textbf{GPT}} &
    \makecell{\textbf{GPT}\\\textbf{4}} &
    \makecell{\textbf{LLaMA}\\\textbf{2-7B}} &
    \makecell{\textbf{Lince-}\\ \textbf{zero}} &
    \makecell{\textbf{Falcon-}\\ \textbf{7B}} &
    \makecell{\textbf{Bloomz-}\\ \textbf{7B1-}\\ \textbf{mt}} &
    \makecell{\textbf{FinMA-}\\ \textbf{7B-}\\ \textbf{full}} &
    \makecell{\textbf{FinMA-}\\ \textbf{30B-}\\ 
    \textbf{nlp}} &
    \makecell{\textbf{FinMA-}\\ \textbf{ES-}\\ \textbf{Bilingual}} &
    \makecell{\textbf{FinMA-}\\ \textbf{ES-}\\ \textbf{Spanish}} 
    \\\midrule
%    \multirow{2}{*}{FPB}
%			&F1&0.78\textcolor{black}{*}&0.78\textcolor{black}{*}&0.51\textcolor{black}{*}&\textcolor{black}{0.29}&\textbf{0.94}&\textcolor{black}{0.03}&\textbf{0.94}&0.88\\\cline{2-10}
%			&Acc&0.78\textcolor{black}{*}&0.76\textcolor{black}{*}&-&\textcolor{black}{0.26}&\textbf{0.94}&\textcolor{black}{0.12}&\textbf{0.94}&0.87\\\hline

\multirow{2}{*}{MultiFin}
& Acc & 0.48 & 0.6 & 0.23 & 0.22 & 0.05 & 0.23 & 0.25 & 0.21 & \textbf{0.99} & 0.91 \\\cline{2-12}
& F1  & 0.47 & 0.6 & 0.11 & 0.1 & 0.07 & 0.16 & 0.27 & 0.19 & \textbf{0.99} & 0.91 \\\hline

\multirow{2}{*}{EFP}
& Acc & 0.30 & 0.27 & 0.27 & 0.27 & 0.27 & 0.38 & 0.35 & 0.38 & \textbf{0.84} & 0.65\\\cline{2-12}
& F1  & 0.26 & 0.19 & 0.12 & 0.12 & 0.12 & 0.38 & 0.21 & 0.29 & \textbf{0.83} & 0.66\\\hline

\multirow{2}{*}{EFPA}
& Acc & 0.31 & 0.34 & 0.26 & 0.25 & 0.26 & 0.51 & 0.35 & 0.34 & \textbf{0.99} & 0.86\\\cline{2-12}
& F1  & 0.25 & 0.27 & 0.1 & 0.1 & 0.10 & 0.52 & 0.21 & 0.26 & \textbf{0.99} & 0.85\\\hline

\multirow{3}{*}{FNS-2023}
& rouge1 & 0.02 & \textbf{0.19} & 0 & 0 & 0 & 0 & 0.01 & 0 & 0 & 0 \\\cline{2-12}
& rouge2 & 0.04 & \textbf{0.06} & 0 & 0 & 0 & 0 & 0 & 0 & 0 & 0 & \\\cline{2-12}
& rougeL & 0.12 & \textbf{0.13} & 0 & 0 & 0 & 0 & 0 & 0 & 0 & 0 & \\\hline

\multirow{2}{*}{TSA}
& Acc & 0.21 & 0.47 & 0.07 & 0.32 & 0.06 & 0.22 & 0.04 & 0.67 & 0.85 & \textbf{0.86}\\\cline{2-12}
& F1  & 0.24 & 0.56 & 0.04 & 0.36 & 0.10 & 0.32 & 0.07 & 0.76 & \textbf{0.86} & 0.86\\\hline

\multirow{2}{*}{FinanceES}
& Acc & 0.13 & 0.15 & 0.14 & \textbf{0.39} & 0.15 & 0.03 & 0.02 & 0.03 & 0.11 & 0.13\\\cline{2-12}
& F1  & 0.08 & 0.09 & 0.13 & \textbf{0.29} & 0.18 & 0.04 & 0.03 &  0.06 & 0.11 & 0.13\\\hline
\hline

\multirow{2}{*}{FPB}
& Acc & 0.78 & 0.76 & 0.68 & 0.51 & 0.64 & 0.39 & 0.87 & \textbf{0.87} & 0.83 & 0.73\\\cline{2-12}
& F1 & 0.78 & 0.78 & 0.65 & 0.52 & 0.64 & 0.23 & 0.87 & \textbf{0.88} & 0.83 & 0.73\\\hline

FiQA-SA
& F1 & 0.6 & 0.8 & 0.77 & 0.82 & 0.77 & 0.77 & 0.79 & \textbf{0.87} & 0.85 & 0.83\\\hline

Headline
& AvgF1 & 0.77 & 0.86 & 0.72 & 0.81 & 0.45 & 0.71 & 0.97  & \textbf{0.97} & 0.96 & 0.58\\\hline

NER
& EntityF1 & 0.77 & \textbf{0.83} & 0 & 0 & 0 & 0 & 0.69 & 0.62 & 0.65 & 0.01\\\hline

FinQA
& EmAcc & 0.58 & \textbf{0.63} & 0 & 0 & 0.002 & 0 & 0.04 & 0.11 & 0.05 & 0\\\hline

ConvFinQA
& EmAcc & 0.60 & \textbf{0.76} & 0 & 0 & 0 & 0 & 0.20 & 0.40 & 0 & 0\\\hline

\multirow{2}{*}{BigData22}
& Acc & 0.53 & 0.54 & 0.51 & 0.55 & 0.55 & 0.55 & 0.49 & 0.47 & 0.48 & \textbf{0.57}\\\cline{2-12}
& MCC & -0.025 & 0.03 & 0.030& 0.000 & 0.000 & -0.007 & 0.010 & 0.040 & 0.100 & \textbf{0.110}\\\hline

\multirow{2}{*}{ACL18}
& Acc & 0.50 & 0.52 & 0.51 & 0.47 & 0.51 & 0.50 & \textbf{0.56} & 0.49 & 0.49 & 0.50\\\cline{2-12}
& MCC & 0.005 & 0.020 & 0.010 & -0.060 & -0.004 & -0.040 & \textbf{0.100} & 0.000 & -0.080 & -0.010\\\hline

\multirow{2}{*}{CIKM18}
& Acc & 0.55 & \textbf{0.57} & 0.47 & 0.43 & 0.44 & 0.55 & 0.53 & 0.43 & 0.42 & 0.55\\\cline{2-12}
& MCC & 0.005 & 0.020 & -0.070 & 0.010 & -0.010 & -0.050 & \textbf{0.100} & 0.000 & -0.040 & -0.040 \\\hline
\hline

\multirow{2}{*}{FinerOrd}
& EntityF1 & 0.28 & \textbf{0.77} & 0 & 0 & 0 & 0 & 0 & 0 & 0 & 0\\\cline{2-12}
& F1 & 0.08 & \textbf{0.78} & 0 & 0 & 0 & 0 & 0 & 0 & 0 & 0\\\hline

\multirow{3}{*}{ECTSum}
& rouge1 & 0 & 0 & 0 & 0 & 0 & 0 & 0 & 0 & 0 & 0  \\\cline{2-12}
& rouge2 & 0 & 0 & 0 & 0 & 0 & 0 & 0 & 0 & 0 & 0\\\cline{2-12}
& rougeL & 0 & 0 & 0 & 0 & 0 & 0 & 0 & 0 & 0 & 0\\\hline

\multirow{3}{*}{EDTSum}
& rouge1 & 0.17 & 0.2 & 0.13 & 0.07 & 0.15 & 0.12 & 0.13 & 0.17 & 0.15 & \textbf{0.26}\\\cline{2-12}
& rouge2 & 0.08 & \textbf{0.19} & 0.06 & 0.03 & 0.06 & 0.06 & 0.06 & 0.08 & 0.07 & 0.14\\\cline{2-12}
& rougeL & 0.13 & 0.15 & 0.12 & 0.07 & 0.13 & 0.12 & 0.10 & 0.14 & 0.14 & \textbf{0.23}\\\hline

\multirow{2}{*}{German}
& Acc & 0.2 & 0.55 & 0.61 & \textbf{0.66} & 0.66 & 0.39 & 0.17 & 0.53 & 0.60 & 0.66 \\\cline{2-12}
& F1 & 0.41 & 0.513 & \textbf{0.60} & 0.52 & 0.52 & 0.40 & 0.17 & 0.53 & 0.60 & 0.52 \\\hline

\multirow{2}{*}{Australian}
& Acc & 0.41 & \textbf{0.74} & 0.43 & 0.43 & 0.47 & 0.57 & 0.41 & 0.46 & 0.72 & 0.56 \\\cline{2-12}
& F1 & 0.26 & \textbf{0.75} & 0.26 & 0.26 & 0.26 & 0.41 & 0.41 & 0.46 & 0.71 & 0.51 \\\hline

\multirow{2}{*}{FOMC}
& Acc & 0.6 & \textbf{0.69} & 0.50 & 0.33 & 0.30 & 0.30 & 0.46 & 0.43 & 0.55 & 0.50 \\\cline{2-12}
& F1 & 0.64 & \textbf{0.71} & 0.35 & 0.28 & 0.30 & 0.20 & 0.49 & 0.53 & 0.49 & 0.46 \\

    \bottomrule
\end{tabular}
}
	\end{table*}
 
\subsection{Results}
\label{sec:results}
\subsubsection{Overall Performance}
Table \ref{tab:per} and Figure \ref{fig:radar} presents a detailed comparative performance analysis of our FinMA-ES models against other leading large language models (LLMs) on the FLARE-ES benchmark. In the realm of Spanish financial tasks, the FinMA-ES-Bilingual and FinMA-ES-Spanish models, both with a 7-billion parameter count, demonstrate superior performance against other state-of-the-art (SOTA) models, including significantly larger models like GPT-4, in four out of six datasets. These datasets include MultiFin, EFP, EFPA, and TSA. The standout performance of our models on these tasks highlights the substantial impact of instruction fine-tuning, which is specifically tailored to enhance the models' understanding and generation of Spanish financial language nuances.

Despite the strong results in most datasets, the FinanceES dataset presents a more competitive scenario, where our models perform on par with SOTA LLMs like GPT-4. However, they do not outperform Lince-zero, which benefits from instruction tuning with a large corpus of Spanish data. This suggests that while our models are highly competitive, there is a unique advantage inherent to models fine-tuned on extensive general domain Spanish data, indicating room for further optimization in future model iterations.
The FNS-2023 dataset, focusing on text summarization, reveals a universal challenge for most models, with only GPT-4 achieving notable success. This might indicate an area where specialized training or model architectures are required to handle the complexities of summarization tasks effectively.
Our models exhibit their robustness within the English datasets, achieving the best results in the FinanceES dataset and comparable performance to SOTA models such as FinMA-30B-nlp and GPT-4 in eight others.

\subsubsection{Generalization Ability.} 
Notably, in three of the six leave-out English datasets—German, Australian, and FOMC—our models emerge as the top performers among all open-source LLMs. Moreover, they maintain competitive performance with renowned models like GPT-4 and ChatGPT in four datasets. This underscores the adaptability and generalization capacity of our models across diverse financial contexts and languages.
For the FinerORD and ECTSum datasets, we implemented a complex prompt design inspired by the Pixiu paper~\cite{xie2023pixiu} which involves direct generation of label sequences. This advanced methodological approach did not yield the desired performance for any model except GPT-4 in the FinerOrd dataset, and none of the models managed to effectively tackle the ECTSum dataset. These results underscore the challenges in generating accurate label sequences and point towards the need for more innovative approaches or specialized training to overcome these hurdles.

\subsubsection{Language Disparity.} 
Table \ref{tab:per} elucidates a marked language disparity when evaluating the proficiency of existing LLMs, including the renowned GPT-4, across Spanish and English financial tasks. This table starkly underscores the challenges LLMs face with Spanish financial tasks, underscoring a considerable performance gap that persists despite advancements in the field.
The data reveals that while these LLMs are adept at handling English financial tasks, they falter significantly when it comes to Spanish. 
The underperformance of these models in Spanish is indicative of the fact that while LLMs have made significant strides, their proficiency is unevenly distributed across languages, with a clear bias towards English.
The performance of our FinMA-ES models shows a significant improvement over other LLMs in Spanish tasks, which can be attributed to the instruction tuning performed with datasets in the target language. 
This suggests that, to address the language disparity in LLMs effectively, a dedicated effort to develop and fine-tune models with a rich and diverse dataset in the target language is essential. 

\subsubsection{Ablation study.}
The ablation study focuses on comparing the performance of monolingual and bilingual models across different datasets. From Table \ref{tab:per}, we can see the FinMA-ES-Bilingual model, which leverages both Spanish and English data, exhibits a distinct advantage over its monolingual counterpart, the FinMA-ES-Spanish model. This superior performance is evidenced in three out of six datasets focused on Spanish financial tasks. The bilingual model's proficiency suggests that exposure to English domain-specific datasets not only reinforces but also enhances its performance in Spanish financial contexts.
Further analysis shows that the FinMA-ES-Bilingual model outperforms the monolingual model in six out of nine English datasets and in half of the leave-out English datasets. The underlying reason for this better performance is attributed to the bilingual model's integration of additional English financial instruction tuning data, which is not utilized by the FinMA-ES-Spanish model. This strategic use of cross-lingual data emphasizes the importance of diverse linguistic training in the development of more adaptable and proficient LLMs.

A pivotal aspect of this study involves the comparison between the FinMA-ES-Spanish model and its foundational model, LLamA2 13B. Intriguingly, the FinMA-ES-Spanish model surpasses LLamA2 13B across all English datasets. This finding is particularly enlightening as it illustrates that fine-tuning LLMs with low-resource language data does not hinder but actually enhances their performance in high-resource language datasets within the financial domain.

Moreover, a comparative review of Lince-zero and Falcon-7B reveals that Lince-zero, which is fine-tuned on a broader set of general domain Spanish data, surpasses Falcon-7B across nearly all Spanish datasets. This reinforces the hypothesis that general domain data in Spanish can significantly enhance model performance on Spanish financial tasks. However, the scenario reverses when it comes to English datasets, where Falcon-7B tends to perform better. This dichotomy underscores a potential trade-off, suggesting that while general domain Spanish data is beneficial for Spanish task performance, it may inadvertently impair the model's effectiveness in English financial tasks.

\section{Conclusion}
This paper addresses the linguistic disparity in the financial area by introducing the first bilingual LLMs framework for Spanish and English. Through the meticulous curation of over 144K bilingual instruction samples and the development of FinMA-ES, a model fine-tuned for bilingual financial analysis, we bridge a crucial gap in the field. Our efforts culminate in the FLARE-ES benchmark, a novel benchmark for comprehensive cross-lingual evaluations, which exposes significant performance gaps and biases in current LLMs. Notably, FinMA-ES demonstrates superior performance over existing SOTA LLMs, including GPT-4, in Spanish financial tasks by effectively leveraging strategic instruction tuning and a diverse dataset for cross-linguistic transfer. By releasing our datasets, models, and benchmarks, we seek to encourage further exploration into complex bilingual scenarios, aiming for more inclusive and effective financial NLP solutions. Looking forward, we aim to further refine our models and evaluation methods, exploring broader applications and improving their performance on complex bilingual scenarios.
 
\section{Limitations}
	\label{sec:limitation}
	Despite the positive contributions of this study, we recognize the following limitations: 1) \textbf{Parameter Restriction}: FinMA-ES is developed with a cap of 7B parameters, a constraint dictated by our available computational resources, which has implications for its depth of training and overall efficacy. 2) \textbf{Evaluation Benchmark Diversity}: The model demonstrates a limited range in its evaluation benchmarks, particularly affecting its capability in tasks like financial summarization. 3) \textbf{Scope of Application}: The specific design and instructional approach of FinMA-ES might limit its applicability across varied bilingual scenarios. 4) \textbf{Ethical and Practical Concerns}: We must consider the potential for negative outcomes, such as disseminating inaccurate financial information or improper market influence. Therefore, we recommend utilizing FinMA-ES primarily for scholarly research, mindful of these ethical aspects. 

\newpage
\bibliographystyle{ACM-Reference-Format}
\bibliography{custom}

\appendix
\section{Instructions}
\label{sec:ins}
\begin{table*}[htbp]
		\centering
		\scriptsize
  	\caption{The example prompts for all Spanish datasets along with their corresponding prompts in Spanish and English translations.
    MultiFin is a classification task and includes article headlines as its text data type. Also, FNS-2023 is a text summarization task that includes annual reports text data type. EFP and EFPA are both question-answering tasks from financial exams in Spanish. TSA  and FinanceES are both sentiment analysis tasks with \textcolor{blue}{\{category\}}: negative, positive, neutral in English and \textcolor{blue}{\{category\}}:negativo, positivo, neutral in Spnaish. % only spanish
   %FiQA-SA has two types of text, including news headlines and tweets. We will fill the detailed text type into \textcolor{blue}{\{category\}} for each data sample. For stock movement prediction data such as BigData22, we will fill \textcolor{blue}{\{tid\}} and \textcolor{blue}{\{point\}} with the detailed stock name and time from each data sample.
   }
		\label{tab:prompt}
        \resizebox{\textwidth}{!}{
        \begin{tabular}{lll}
			\toprule
			\textbf{Dataset} & \textbf{Spanish Prompts} &  \textbf{English Translations}\\
			\midrule
			FinanceES & \makecell[l]{``¿Cuál es el sentimiento de esta oración? Responde solo negativo, positivo o neutral.\\ \textcolor{blue}{\{category\}}:positivo, negativo, or neutral?"} & \makecell[l]{What is the sentiment of this sentence?\\ Answer only negative, positive or neutral.}\\
			\midrule
            TSA & \makecell[l]{``¿Cuál es el sentimiento de esta oración? Responde solo negativo, positivo o neutral.\\ \textcolor{blue}{\{category\}}:positivo, negativo, or neutral?"} & \makecell[l]{What is the sentiment of this sentence?\\ Answer only negative, positive or neutral.}\\
                \midrule
			FNS-2023 & \makecell[l]{``Por favor, lea el texto con atención y resuma su contenido de forma breve y precisa."} & \makecell[l]{Please read the text carefully and summarize the content of the text accurately and birefly.} \\
			\midrule
			EFP & \makecell[l]{``Lea cuidadosamente las preguntas y respuestas, \\y elija la que considere apropiada entre las tres opciones A, B y C."} & \makecell[l]{Read the questions and answers carefully,\\ and choose the option that you think is appropriate from the three options A, B and C.}\\
			\midrule
			EFPA & \makecell[l]{``Lea cuidadosamente las preguntas y respuestas, \\y elija la que considere apropiada entre las tres opciones A, B ,C y D. "} & \makecell[l]{Read the questions and answers carefully,\\ and choose the option that you think is appropriate from the three options A,B, C and D.}\\
			\midrule
            MultiFin & \makecell[l]{``Lee el texto cuidadosamente y elige la etiqueta adecuada para el texto de las etiquetas de \\ 'Negocios y Gestión', 'Finanzas', \\'Gobierno y Control', 'Industria', 'Impuestos y Contabilidad', 'Tecnología' \\ \textcolor{blue}{\{category\}}:'Negocios y Gestión', 'Finanzas', 'Gobierno y Control', 'Industria', \\'Impuestos y Contabilidad', 'Tecnología' ".} & \makecell[l]{Read the text carefully and choose one appropriate label for the text from the labels of \\ 'Business and Management', 'Finance', 'Government and Controls', 'Industry', \\'Tax and Accounting', 'Technology'.}\\
            
            %BigData22 & \makecell[l]{``Analiza la información y las publicaciones en redes sociales para determinar si el precio de cierre \\de \textcolor{blue}{\{tid\}} subirá o bajará en \textcolor{blue}{\{point\}}. \\Por favor, responde con Sube o Baja.``} & \makecell[l]{Contemplate the data and tweets to guess whether the closing price of \textcolor{blue}{\{tid\}} will surge or \\decline at \textcolor{blue}{\{point\}}.\\ Please declare with either Rise or Fall."} \\

			\bottomrule
		\end{tabular}
        }
		%\caption{The example prompt for each dataset in both English and Spanish. MultiFin is a classification and include article headlines text type. Also, FNS-2023 is a text summarizaiton task which include annual reports text type. EFP and EFPA are both question answering task from financial exams in spanish. TSA  and FInanceES are both sentiment analysis task with \textcolor{blue}{\{category\}}: negative, positive, neutral in English and \textcolor{blue}{\{category\}}:negativo, positivo, neutral in Spnaish. For stock movement prediction data such as BigData22, we will fill \textcolor{blue}{\{tid\}} and \textcolor{blue}{\{point\}} with the detailed stock name and time from each data sample.  %FiQA-SA has two types of text, including news headlines and tweets. We will fill the detailed text type into \textcolor{blue}{\{category\}} for each data sample. For stock movement prediction data such as BigData22, we will fill \textcolor{blue}{\{tid\}} and \textcolor{blue}{\{point\}} with the detailed stock name and time from each data sample.}
		%\label{tab:prompt}
		
	\end{table*}

\end{document}